\documentclass{article}
\usepackage{spconf,amssymb,amsmath,epsfig}
\usepackage{microtype}
\usepackage{graphicx}
\usepackage{multirow}
\usepackage{booktabs}
\usepackage{amssymb,amsmath,graphicx}
\usepackage{caption}
\usepackage{subcaption}
\usepackage[export]{adjustbox}
\usepackage{moredefs} 
\usepackage{color}

\newcommand{\red}[1]{{\color{red} \textbf{#1}}}
\defcommand{\vec}[1]{\mathbf{#1}} 

\def\vx{\mathbf{x}}
\def\vh{\mathbf{h}}
\def\vy{\mathbf{y}}

\def\eps{\texttt{<epsilon>}}

\title{Improving the Performance of Online Neural Transducer Models}
\name{Tara N. Sainath, Chung-Cheng Chiu, Rohit Prabhavalkar, Anjuli Kannan, 
\secondlinename{Yonghui Wu, Patrick Nguyen, Zhifeng Chen}
\address{Google, Inc., USA \\
\fontsize{9}{9}\selectfont\ttfamily\upshape
\{tsainath,chungchengc,prabhavalkar,anjuli,yonghui,drpng,zhifengc\}@google.com}}

\begin{document}
\maketitle
\ninept
\begin{abstract}
  Having a sequence-to-sequence model which can operate in an online fashion is important for streaming applications such as Voice Search.
  Neural transducer is a streaming sequence-to-sequence model, but has shown a significant degradation in performance compared to non-streaming models such as Listen, Attend and Spell (LAS). In this paper, we present various improvements to NT. Specifically, we look at increasing the window over which NT computes attention, mainly by looking backwards in time so the model still remains online. In addition, we explore initializing a NT model from a LAS-trained model so that it is guided with a better alignment. Finally, we explore including stronger language models such as using wordpiece models, and applying an external LM during the beam search. On a Voice Search task, we find with these improvements we can get NT to match the performance of LAS.
\end{abstract}

\section{Introduction \label{sec:introduction}}

Sequence-to-sequence models have become popular in the automatic speech recognition (ASR) community~\cite{Jan15,Chan15,Jaitly16,RohitAnal17}, as they allow for one neural network to jointly learn an acoutic, pronunciation and language model, greatly simplifying the ASR pipeline. In this paper, we focus on attention-based sequence-to-sequence models, as our previous study~\cite{RohitSeq17} showed these models performed better than alternatives such as Connectionist Temporal Classification (CTC)~\cite{Graves06} and Recurrent Neural Network Transducer (RNN-T)~\cite{Graves12}.

Attention-based models consist of three modules. First, an \emph{encoder}, represented by a multi-layer recurrent neural network (RNN), models the acoustics. Second, a \emph{decoder}, which consists of multiple RNN layers, predicts the output sub-word unit sequence. Finally, an \emph{attention} layer selects frames in the encoder representation that the decoder should attend to when predicting each sub-word unit.

Attention-based models, such as Listen, Attend and Spell (LAS) have typically been explored in ``full-sequence'' mode, meaning attention is computed by seeing the entire input sequence~\cite{Chan15,RohitAnal17}. Thus, during inference, the model can produce the first output token only after all input speech frames have been consumed. While such a mode of operation might be suitable for many applications, these models cannot be used for ``streaming" speech recognition, such as voice search, where the output text should be generated as soon as possible after words are spoken \cite{Matt17}. 

Recently, neural transducer (NT)~\cite{Jaitly16} was proposed as a limited-sequence streaming attention-based model, which consumes a fixed number of input frames (a chunk), and outputs a variable number of labels before it consumes the next chunk. While the model is attractive for streaming applications, in previous work NT showed a large degradation over other online sequence-to-sequence models such as RNN-T \cite{Battenberg17} and full-sequence unidirectional attention-based models~\cite{Jaitly16, RohitAnal17}, particularly as the chunk-size was decreased~\cite{RohitAnal17}.

In the present work, we study various improvments to the streaming NT model\footnote{In this work, we consider streaming to mean the system has a maximum allowable delay of 300ms, which is considered reasonable \cite{Matt17}.}  -- both in terms of model structure, as well as in the training procedure -- that are aimed at improving its performance to be as close as possible to the non-streaming full-sequence unidirectional LAS model, which serves as an upper-bound of sorts. Specifically, we allow attention in NT to be computed \emph{looking back many previous chunks}, as this does not introduce additional latency. Further, we find that allowing the model to look-ahead by 5 frames is extremely beneficial. Finally, we allow NT to be initialized from a pre-trained LAS model, which we find is a more effective strategy than having the model learn from scratch.

Our NT experiments are conducted on a 12,500 hour Voice Search task. We find that with look-back and look-ahead, NT is more than 20\% relative worse than LAS in terms of word error rate (WER). However, we find that by pretraining with LAS, we can get NT with a chunk size of 10 (450ms latency) to match the performance of LAS, but a chunk size of 5 (300ms latency) still degrages by 3\% relative.

Our analysis of the NT model indicates that many of the errors made compared to LAS are language modeling (LM) errors. Thus, we explore various ideas to incorporate a stronger language model (LM) into NT, to allow us to reduce the chunk size. This includes exploring incorporating an LM from the encoder side via multi-head attention, training the NT model with word pieces~\cite{Schuster2012} to get a stronger LM into the decoder~\cite{Rao17} and also explicitly incorporating an external LM via shallow fusion~\cite{Jan17}. We find that our best performing NT system with a chunk size of 5 (300 ms latency) only degrades performance by 1\% relative to an unidirectional LAS system.


\section{Original Neural Transducer Algorithm}\label{sec:nt}
\begin{figure}
\centering
  \includegraphics[scale=0.35]{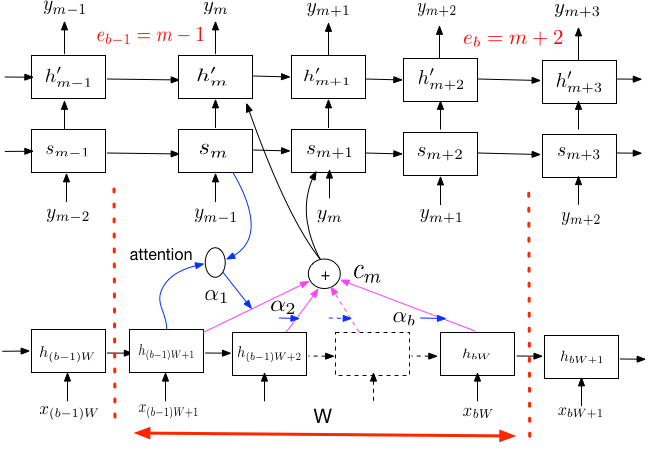}\\
  \caption{Neural Transducer Attention Model.}
  \label{fig:nt}
  \vspace{-0.2in}
\end{figure}

In this section, we describe the basic NT model introduced in Jaitly et
al.~\cite{Jaitly16}. which is shown in Figure \ref{fig:nt}. Given an input sequence of frame-level features (e.g., log-mel-filterbank
energies), $\vx=\{x_1, x_2, \ldots, x_T\}$, and an output sequence of sub-word
units (e.g., graphemes, or phonemes) $\vy=\{y_1, y_2, \ldots y_N\}$, attention models
assume that the probability distribution of each sub-word 
unit is conditioned on the previous history of sub-word unit predictions,
$y_{<i}$, and the input signal. Full-sequence attention models, such as LAS \cite{Chan15} compute
the probability of the output prediction $y_{<i}$ for each step $i$ given the entire input acoustic sequence
$\vx$, making it unsuitable for streaming recognition applications.
The Neural Transducer (NT) model~\cite{Jaitly16} is a limited-sequence attention
model that addresses this issue by limiting attention to fixed-size blocks
of the encoder space. 

Given the input sequence, $\vx$, of length $T$, and a block size of length $W$, the
input sequence is divided equally into blocks of length
$B=\lceil{\frac{T}{W}}\rceil$, except for the last block which might contain
fewer than $B$ frames.
The NT model examines each block in turn, starting with the left-most block (i.e., the
earliest frames).
In this model, attention is only computed over the frames in each block.
Within a block, the NT model produces a sequence of $k$ outputs, $y_i, \ldots,
y_{i+k}$; it is found to be useful to limit the maximum number of outputs that
can be produced within a block to $M$ symbols, so that $0\leq k\le M$.
Once it has produced all of the required labels within a block, the model
outputs an \eps\texttt{} symbol, which signifies the end of block processing.
The model then proceeds to compute attention over the next block, and
so on, until all blocks have been processed.
The \eps\texttt{} symbol is analogous to the \emph{blank} symbol in
connectionist temporal classification (CTC)~\cite{Graves06}.
In particular, we note that a block must output a minimum of one symbol
(\eps\texttt{}), before proceeding to the next block.

The model computes $P(y_{1, \ldots, (S+B)} | \vx_{1 \ldots T})$, which
outputs a sequence which is length $B$ longer than the LAS model since the model
must produce an \eps\texttt{} at every block. Within each block $b \in B$, the model
computes the following probability in Equation~\ref{eq:nt}, where $y_{e_b}=\eps$
is the symbol at the end of each block.
In other words, the prediction $y_i$ at the current step, $i$, is based on the
previous predictions $y_{1\ldots {e_{(i-1)}}}$, similar to LAS, but in this case
using acoustic evidence only up to the current block, $\vx_{1\ldots bW}$:
\begin{multline}
  P(y_{(e_{b-1}+1)\ldots e_b} | \vx_{1\ldots bW},y_{1\ldots {e_{b-1}}}) = \\
  \prod_{i=e_{(b-1)}+1}^{e_b} P(y_i |  \vx_{1\ldots bW},y_{1\ldots {e_{(i-1)}}})
  \label{eq:nt}
\end{multline}

Like LAS, NT also consists of  a \emph{listener}, an \emph{attender} and a
\emph{speller} to define a probability distribution over the next
sub-word unit conditioned on the acoustics and the sequence of previous
predictions. The listener module of the NT computes an encoding vector in the current block only:
\begin{equation}
	\vh_{(b-1)W+1\ldots bW} = \text{Listen}(\vx_{(b-1)W+1\ldots bW})
  \label{eq:listen_nt}
\end{equation}
\noindent which is implemented as a \textit{unidirectional} RNN.

The goal of the attender and speller is to take the output of the listener
(i.e., $\vh$) and produce a probability distribution over sub-word units. The attention and
speller modules operate similar to LAS, but only work on the partial output,
$\vh_{1\ldots bW}$, of the encoder up until the current block. We refer the reader
to \cite{RohitAnal17} for more details about the attention and speller.

\section{Improving Performance of Basic Neural Transducer Algorithm}
In this section, we describe various improvements to the basic algorithm and
model described in the previous section.

\subsection{Training grapheme-based models using Word Alignments}

Training with NT requires knowing which sub-word units occur in each chunk, and thus an alignment is needed. Our previous work with NT \cite{RohitAnal17} used context-independent phonemes, for which an alignment was available. In this work, we train our model with graphemes, which does not have an alignment. However, we have a word level alignment and we use this information to emit all graphemes in the chunk corresponding to when a word has finished. 

\subsection{Extending Attention Range}
In the original NT paper, attention was computed by only looking at encoder features
in the current block $b$, as shown in Equation \ref{eq:listen_nt}. However, as shown in \cite{RohitAnal17},
making the attention window longer allows NT to approach the performance of LAS, but at the cost of removing
the online nature of the task. However, we can still maintain
a streaming, online system by computing attention by looking back over $k$ previously blocks. This is particularly important because we emit graphemes at word boundaries. Furthermore, similar to our streaming systems \cite{Golan16}, we allow a lookahead of 150 ms (5 30-ms frames) between the input frames
and the output prediction. With these changes, the listener is now shown by Equation \ref{eq:listen_nt_streaming}.

\begin{equation}
        \vh_{(b-1)W+1\ldots bW} = \text{Listen}(\vx_{(b-k)W+1\ldots bW + 5})
  \label{eq:listen_nt_streaming}
\end{equation}

\subsection{Pre-training with LAS}

Attention-based models learn an alignment (represented via an attention vector), jointly with the acoustic model (encoder) and language model (decoder). One 
hypothesis we have for NT lagging behind LAS is that during training, the attention mechanism is limited in the window over which it can compute attention. This problem is exacerbated by the fact that we emit graphemes only at word boundaries. 

However, we can see from attention plots in LAS \cite{Chan15} that once the attention mechanism is learned, it appears to be fairly monotonic. Since NT and LAS are parameterized exactly the same (except for an extra \eps\texttt{} output target), we can train a LAS model with this extra target (which is ignored as it does not appear in the LAS target sequence) and used it to initialize NT. Our intuition is that since LAS learns a good attention mechanism that is relatively monotonic, it can be used to initialize NT and NT will not take a large hit in accuracy compared to LAS.

\subsection{Incorporating a Stronger Language Model}

As we make the chunk-size smaller, looking at the errors it appears that most of the errors are due to language modeling errors.
Therefore, we explore if we can incorporate a stronger LM into the decoding and/or training process. 



\subsubsection{Wordpiece Models}

To increase the memory and linguistic span of the decoder, we emit wordpieces instead of graphemes \cite{Yonghui16}. In this approach, words a broken up, deterministically, into sub-word units, called wordpieces.  For instance, the phrase ``Jet makers feud'' can be broken up into (``\_J'' , ``et'', ``\_makers'', ``\_fe'', ``ud'') some words may be broken down into sub-units while common words (``makers'') are modeled as a single unit. Wordpieces are position-dependent, so we mark the beginning of each word with a special marker ``\_''. The wordpiece inventory is trained to maximize the likelihood of the training text. Wordpieces achieve a balance between the flexibility of characters and efficiency of words.

Sequence-to-sequence models that predict wordpieces have been successful in both machine translation \cite{Yonghui16} and speech \cite{Rao17,lsd2017iclr}. Since these models are trained to predict wordpieces, rather than graphemes, a much stronger decoder LM is used. We hypothesize that by predicting wordpieces, we can reduce chunk size as well with NT.

\subsubsection{Incorporating external LM}
Language models have been successfully incorporated into sequence-to-sequence models to guide the beam search to output a more likely set of candidates \cite{Chorowski17,ColdFusion17}. In this work, we explore if incorporating an external LM into the beam search can aid NT. Following a similar approach to \cite{Chorowski17,Anjuli18}, we look at doing a log-linear interpolation between the LAS model and an FST-based LM trained to go from graphemes to words at each step of the beam search, also known as shallow fusion \cite{ColdFusion17}.  In this equation $ p(\vy | \vx)$ is the score from the LAS model, which is combined with a score coming from an external LM $p_{LM}(\vx)$ weighted by an LM weight $\lambda$, and a \texttt{coverage} term to promote longer transcripts \cite{Chorowski17} and weighted by $\eta$.

\begin{equation}
  \vy^* = \arg \min_y - \log p(\vy | \vx) - \lambda \log p_{LM}(\vx) - \eta\texttt{coverage}
  \label{eq:lm}
  \vspace{-0.1in}
\end{equation}

\section{Experimental Details \label{sec:experiments}}

Our experiments are conducted on a $\sim$12,500 hour training set consisting of
15 million English utterances.
The training utterances are anonymized and hand-transcribed, and are
representative of Google's voice search traffic.
This data set is created by artificially corrupting clean utterances using a
room simulator, adding varying degrees of noise and reverberation such that the
overall SNR is between 0dB and 30dB, with an average SNR of 12dB \cite{Chanwoo17}. The noise sources are from YouTube and daily life noisy environmental
recordings. We report results on a set of $\sim$14,800 anonymized, hand-transcribed Voice Search utterances extracted from Google traffic.

All experiments use 80-dimensional log-mel features, computed with a 25-ms
window and shifted every 10ms. Similar to~\cite{Golan16,Hasim15}, at the current frame, $t$, these features are stacked with 2 frames to the left and downsampled to a 30ms frame rate. The encoder network architecture consists of 5 unidirectional long short-term memory~\cite{HochreiterSchmidhuber97} (LSTM) layers, with the size specified in the results section.  Additive attention is used for all experiments \cite{Bahdanau14}. The decoder network is a 2 layer LSTM with 1,024 hidden units per layer. All networks are trained to predict 74 graphemes unless otherwise noted.

All neural networks are trained with the cross-entropy criterion, using
asynchronous stochastic gradient descent (ASGD) optimization~\cite{Dean12} with Adam \cite{KingmaBa15} and are trained using TensorFlow~\cite{AbadiAgarwalBarhamEtAl15}.

\section{Results \label{sec:results}}

\subsection{Getting NT To Work Online}

\subsubsection{Attention Window}

Our first set of experiments analyzes the behavior of NT as we vary the window used to compute attention.
For these experiments, we use an encoder which consists of five layers of 768
uni-directional LSTM cells and a decoder with two
layers of 768 LSTM cells.
As can be seen in Table~\ref{table:nt_first}, when we only allow the NT model to
compute attention within a chunk of size 10, performance is roughly 25\% worse in terms of WER compared to the LAS model which differs only in the window over which attention is computed. Allowing the model to compute attention over the last 20 chunks in addition to the current chunk, however, slightly improves performance of the NT system. Finally, if we allow a 5 frame look-ahead \footnote{It is important to note that the 5 frame look-ahead with a chunk size of 10 is not the same as a 15 frame window, as the 5 frame look ahead is with respect to the end of the chunk boundary, and all other frames used to compute attention occur before the chunk boundary.}, the results improve but NT is still roughly 13\% relative worse compared to LAS.  Based on the results in Table~\ref{table:nt_first}, since the proposed changes improve performance, all future NT results in the paper use a look-back of 20 chunks and a look-ahead of 5 frames.

\begin{table} [h!]
\centering
\begin{tabular}{|c|c|c|c|} \hline
System & Chunk Size & WER \\ \hline
LAS &  - &  11.7  \\ \hline
NT, attention within chunk & 10 & 14.6 \\ \hline
NT, look back  & 10 &  14.4  \\ \hline                                       
+ look ahead & 10 &  13.2  \\ \hline
\end{tabular}
\vspace{-0.1 in}
\caption{WER for NT, Varying Chunks Looked Over}
\vspace{-0.2 in}
\label{table:nt_first}
\end{table}

\subsubsection{Initialization from LAS, Single-head attention}
Next, we analyze the behavior of NT, for both a chunk size of 5 and 10, when we pretrain with LAS. For these experiments, we compare two different encoder/decoder sizes. Table \ref{table:nt_pretrain} shows that when NT is pre-trained with LAS, at a chunk size of 10 (i.e., 450 ms latency) we can match the performance of LAS. However, a chunk size of 5 (300ms latency), which is our requirement for allowed streaming delay, still lags behind LAS by 3\% relative for the larger model.

\begin{table} [h!]
\centering
\begin{tabular}{|c|c|c|c|} \hline
\multirow{2}{*}{System} & \multirow{2}{*}{Chunk} & 5x768 & 5x1024 \\ 
  & & 2x768 & 2x1024 \\ \hline
LAS & - &  11.7 & 9.8  \\ \hline                                          
NT, scratch & 10 &   13.2  &  11.1  \\ \hline                                       
NT, pretrained & 10 & \textbf{11.4} &  \textbf{9.9}  \\ \hline                                              
NT, scratch & 5 & -  & 14.5   \\ \hline                                       
NT, pretrained & 5 &  -  & 10.1   \\ \hline    
\end{tabular}
\vspace{-0.1 in}
\caption{WER for NT, Pretrained from LAS}
\vspace{-0.1 in}
\label{table:nt_pretrain}
\end{table}

\subsubsection{Initialization from LAS, Multi-head attention}

\begin{table*} [t!]
\centering
\begin{tabular}{|c|c|c|} \hline
LAS, MHA &  NT-Ch5, MHA & NT-Ch5, MHA, WPM \\ \hline
school closing in & \red{what}     closing    in & school closing in \\ 
parma for tomorrow & parma for     tomorrow & parma for tomorrow\\ \hline
how to multiply two  & how to    multiply two & how to multiply two     \\
numbers with decimals & numbers with  \red{this   most}  & numbers with decimals   \\ \hline
how far is it from albuquerque new &  how far is it from albuquerque new & how far is it from albuquerque new \\ 
mexico to fountain hills arizona & mexico \red{to to} fountain hills arizona & mexico to fountain hills arizona \\ \hline
is under the arm warmer or colder & is  under  the   arm  warmer  or   colder & is under the arm warmer or colder \\
than in mouth temperature & than \red{a   mouse}     temperature &  than in mouth temperature  \\ \hline

\end{tabular}
\vspace{-0.1 in}
\caption{Representative errors made by different systems, indicated in \red{red}.}
\vspace{-0.2 in}
\label{table:nt_errors1}
\end{table*}

Next, we compare the behavior of LAS vs. NT when the system uses multi-head attention \cite{Vaswani17}, which has been shown to give state-of-the-art ASR performance for LAS \cite{CC18}. The MHA model uses a 5x1400 encoder with 4 attention heads, and a 2x1024 decoder. Table \ref{table:nt_mha} shows that the performance of NT does not improve from single to multi-head attention, even though the LAS system does. One hypothesis is that multi-head attention computes attention from multiple points in the encoder space that come after the current prediction, which are ignored by streaming models such at NT. 

\begin{table} [h!]
\centering
\begin{tabular}{|c|c|c|c|} \hline
  \multirow{2}{*}{System} & \multirow{2}{*}{Chunk} & Single Attention - WER & MHA - WER \\
                          &                        & 5x1024,2x1024 & 5x1400,2x1024 \\ \hline
LAS & - & 9.8 & 8.0 \\ \hline
NT & 10 & 9.9 & 9.8 \\ \hline
NT & 5 & 10.1 & 10.3 \\ \hline
\end{tabular}
\vspace{-0.1 in}
\caption{WER for NT with MHA}
\vspace{-0.1 in}
\label{table:nt_mha}
\end{table}

To understand the loss in performance caused by NT compared to LAS, we analyzed sentences where LAS was correct and NT incorrect, denoted in the first two columns of Table \ref{table:nt_errors1} as ``LAS-MHA'' and ``NT-Ch5,MHA''. The table shows that a lot of the NT errors are due to language modeling errors. In the next section, we look at a few simple ways of incorporating an LM into the system.


\subsection{Incorporating the LM}


\subsubsection{Wordpieces}

\begin{figure}
\centering
  \includegraphics[scale=0.40]{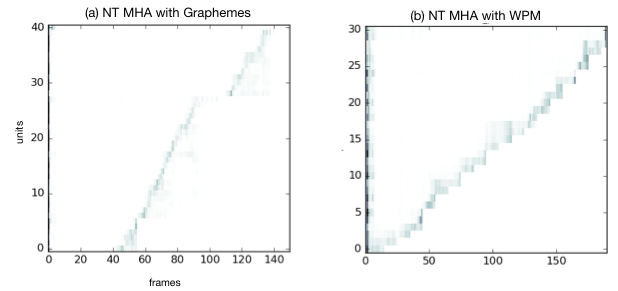} \\
  \caption{Attention Plots for NT-MHA with (a) Graphemes, (b) WPM. The two attention plots also correspond to different utterances.}
  \label{fig:nt_attention}
  \vspace{-0.2in}
\end{figure}

Our next set of results looks at incorporating wordpieces into the LAS and NT models, which provide a stronger LM from the decoder side. For these experiments, we used 32,000 wordpieces. Table \ref{table:nt_wpm_mha} shows that with wordpieces, the NT and LAS models are now much closer compared to graphemes. In addition, there is very little difference between NT with a chunk size of 5 and 10. One hypothesis is that since wordpieces are now longer units, each attention head focused on by neural transducer corresponds to a much longer subword unit (potentially a word) compared to the NT grapheme MHA system. Therefore, the MHA WPM feeds a much stronger set of context vectors to the decoder compared to NT grapheme model. This can also be visually observed by looking at the attention plots for the grapheme vs. wordpiece systems in Figure \ref{fig:nt_attention}. The plot shows that the attention vectors for wordpieces span a much longer left context window compared to graphemes.

\begin{table} [h!]
\centering
\begin{tabular}{|c|c|c|} \hline
System & Chunk &  WER \\ \hline
LAS & - & 8.6 \\ \hline
NT & 10 & 8.6  \\ \hline
NT & 5 & 8.7 \\ \hline
\end{tabular}
\vspace{-0.1 in}
\caption{WER for NT with MHA + WPM}
\vspace{-0.1 in}
\label{table:nt_wpm_mha}
\end{table}

\subsubsection{WPM + LM}

Finally, we investigate incorporating an external LM into the MHA+WPM LAS and NT models. In these experiments, a n-gram FST LM, trained on 32K wordpieces, is used. This LM is trained on 1 billion text queries, a much larger set compared to the 15 million utterances seen by the LAS/NT models.  Table \ref{table:nt_lm} shows the the FST LM does not give any additional improvement for both NT and LAS. It has been observed in \cite{Anjuli18} that the perplexity of a WPM RNN-LM is much lower than a WPM FST. Since the decoder of the LAS and NT models is an RNN-LM, it is possible there is nothing more to gain by incorporating the WPM FST. In the future, we will repeat this with a WPM RNN-LM trained on text data.

\begin{table} [h!]
\centering
\begin{tabular}{|c|c|c|c|} \hline
System & Chunk &  No LM & with LM \\ \hline
LAS & - & 8.6 & 8.6 \\ \hline
NT & 10 & 8.6 &  8.6 \\ \hline
NT & 5 & 8.7 & 8.7 \\ \hline
\end{tabular}
\vspace{-0.1 in}
\caption{WER for NT, Incorporating External LM}
\vspace{-0.1 in}
\label{table:nt_lm}
\end{table}

Finally, it should be noted that after including both WPM and external LM, the last column of Table \ref{table:nt_errors1}, namely ``NT-Ch5,MHA,WPM'' illustrates that many of the previous sentences are now fixed and match the LAS hypothesis. With the proposed LM improvements, NT with a chunk size of 5 has comparable performance to LAS, while meeting the allowable delay of 300ms.

\section{Conclusions \label{sec:conclusions}}
In this paper, we presented various improvements to NT. Specifically, we showed we could improve performance by increasing the attention window and pre-training NT with LAS. With these improvements, a single-head NT model could come very close to the performance of LAS while a multi-head attention NT model still degraded over LAS. By incorporating a stronger LM through wordpieces, multi-head NT could effetively match the performance of LAS.

\section{Acknowledgements}
The authors would like to thank Navdeep Jaitly, Michiel Bacchiani and Gabor Simko for helpful discussions.
\bibliographystyle{IEEEbib}
\bibliography{main}
\end{document}